\documentclass[letterpaper, 10 pt, conference]{ieeeconf}  

\IEEEoverridecommandlockouts                              
\makeatletter
\let\NAT@parse\undefined
\makeatother

\usepackage[dvipsnames]{xcolor}

\newcommand*\linkcolours{ForestGreen}

\usepackage{times}
\usepackage{graphicx}
\usepackage{amssymb}
\usepackage{gensymb}
\usepackage{amsmath}
\usepackage{breakurl}

\usepackage{url,hyperref}
\hypersetup{
colorlinks,
linkcolor=\linkcolours,
citecolor=\linkcolours,
filecolor=\linkcolours,
urlcolor=\linkcolours}

\usepackage{algorithm}
\usepackage{algorithmic}

\usepackage[labelfont={bf},font=small]{caption}
\usepackage[none]{hyphenat}

\usepackage{mathtools, cuted}

\usepackage[noadjust, nobreak]{cite}

\usepackage{tabularx}
\usepackage{amsmath}

\usepackage{float}

\usepackage{pifont}

\newcolumntype{Y}{>{\centering\arraybackslash}X}

\usepackage[]{placeins}

\usepackage{placeins}

\usepackage{tikz}

\usepackage[framemethod=tikz]{mdframed}
\usepackage{diagbox}
\usepackage{afterpage}

\usepackage{stfloats}

\usepackage{atbegshi}
\newcommand{\handlethispage}{}
\newcommand{\discardpagesfromhere}{\let\handlethispage\AtBeginShipoutDiscard}
\newcommand{\keeppagesfromhere}{\let\handlethispage\relax}
\AtBeginShipout{\handlethispage}

\usepackage{comment}
\usepackage{mathtools}
\DeclarePairedDelimiter{\floor}{\lfloor}{\rfloor}
\DeclarePairedDelimiter{\ceil}{\lceil}{\rceil}

\title{\LARGE \bf
Search Algorithms for Mastermind
}

\author{Anthony D. Rhodes\\
Portland State University} 

\begin{document}

\maketitle
\thispagestyle{empty}
\pagestyle{empty}

\begin{abstract}

This paper presents two novel approaches to solving the classic board game mastermind, including a variant of simulated annealing (SA) and a technique we term \textit{maximum expected reduction in consistency} (MERC). In addition, we compare search results for these algorithms to two baseline search methods: a random, uninformed search and the method of minimizing maximum query partition sets as originally developed by both Donald Knuth [3] and Peter Norvig [4].\\

\end{abstract}

\section{INTRODUCTION}

Mastermind is a popular code-breaking two player game originally invented in the 1970s. The gameplay closely resembles the antecedent pen and paper game called "Bulls and Cows", which dates back at least a century. \\
\indent Mastermind consists of three components: a decoding board which includes a dozen or so rows of holes for query pegs, in addition to smaller holes for key pegs; the board also contains a space for the placement of the master code provided by the code-maker. In addition, the game is played with code pegs of different colors placed by the code-breaker (six is the default number of colors, although many variations exist) and key pegs consisting of two colors placed by the code-maker (see Figure 1).\\ 
\indent The code-maker chooses a pattern of four code pegs for the master code. The master code is placed in the four holes covered by the shield, visible to the code-maker but not to the code-breaker. For each turn of the game, the code-breaker attempts to guess the master code with respect to both order and color. Each query is made by placing a row of code pegs on the decoding board adjacent to the last query row. After each guess, the code-maker provides feedback to the code-breaker in the form of key pegs. Between zero and four key pegs are placed next to the query code on the current turn to indicate the fidelity of the current query -- a colored or black key peg connotes a query code peg that is correct in both color and position, whereas a white key peg indicates the existence of a correct color code peg placed in the wrong position. The goal of the game is for the code-maker to determine the master code using a minimal number of queries. \\
\indent Due to its status as a relatively simple, query-based game of incomplete information, mastermind has served as an enduring test-bed for a diverse array of search algorithms in A.I. and related fields. 

\begin{figure}[tbp]
\centering
\includegraphics[width=0.97\columnwidth]{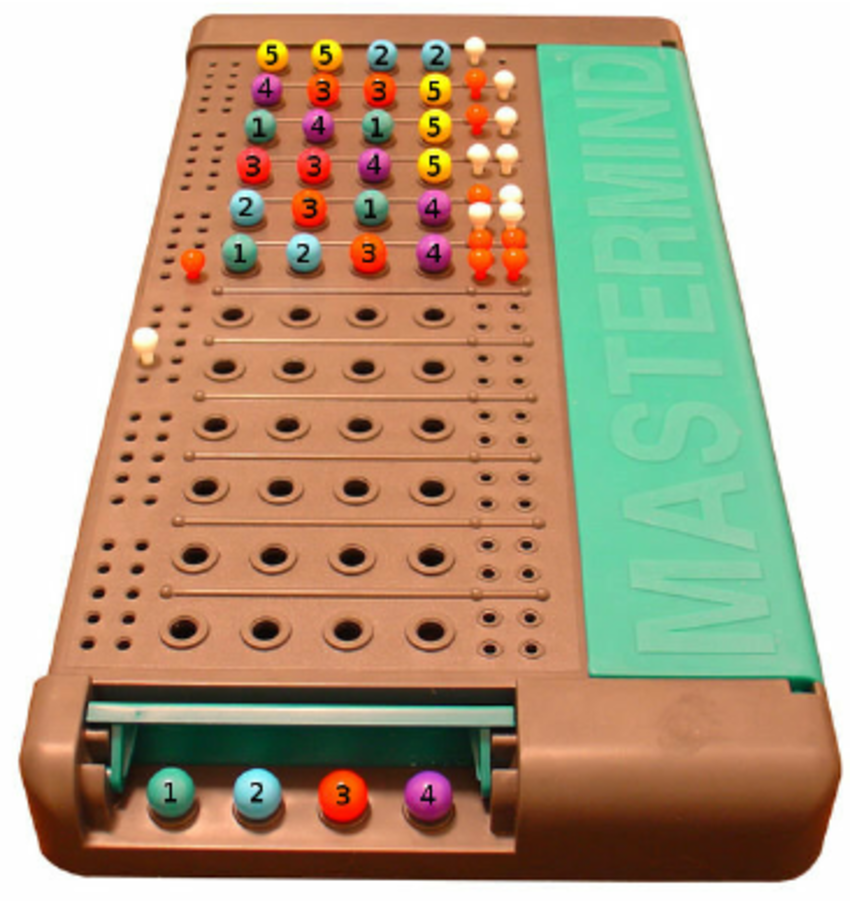}
\caption{ Mastermind game schematic, including decoding board, code pegs and key pegs. A completed game is shown. }
\label{loss_spikes}
\end{figure}
\section{Previous Work and Preliminaries}
Mastermind and its variants have inspired a good deal of research, particularly in the domains of combinatorics and search algorithms. \\
\indent With four pegs and six colors (which we henceforth denote $MM(6,4)$) there are $6^{4}=1296$ possible codes. One of the most essential properties surrounding efficient search in mastermind is the notion of \textit{code consistency}: 
\begin{equation}
\textrm{if } q(m)=q(c) \textrm{ and } q(m)=q(c'), \textrm{ then } c\sim c'
\end{equation}
where above $q(\cdot)$ connotes the query operation, outputting a 2-d key code for a given query code input (e.g. $q(c)=[2,1]$ indicates the query code $c$ generated two black and one white key code responses, respectively, for the given master code); $m$ in equation (1) denotes the master code. In short, code consistency forms an equivalence relation (indicated by $\sim$) over the set of all possible codes. \\
\indent It is not difficult to determine the total number of distinct query partition classes for a generic $MM(c,p)$ game. Notice that all key code combinations, $q(c)=[b,w]$ where $b+w=p$ are possible, $b,w \in \mathbb{Z}_{\ge 0}$ with the exception of [p-1,1]. Thus the total number of distinct query partition classes $Q_{p}$ for $MM(c,p)$ is given by: 
\begin{equation}
Q_{p}=2+\sum_{i=3}^{p+1} i = \frac{(p+2)(p+1)}{2} - 1 = \frac{(p+3)p}{2}
\end{equation}

In particular, for $MM(6,4)$, $Q_{p}=14$; see Table 1 for more details. 

\begin{table}
\centering
\begin{tabular}{|l|c|c|c|c|c|}\hline
\diagbox{b}{w}&
  0 & 1 & 2 & 3 & 4 \\ \hline
0 & [0,0] & [0,1] & [0,2] & [0,3] & [0,4] \\ \hline
1 & [1,0] & [1,1] & [1,2] & [1,3] & X \\ \hline
2 & [2,0] & [2,1] & [2,2] & X & X \\ \hline
3 & [3,0] & X & X & X & X \\ \hline
4 & [4,0] & X & X & X & X \\ \hline
\end{tabular}
\caption{Legal query partition classes for $MM(6,4)$, where $Q_{p}=14.$} 
\end{table}
\indent One may, alternatively, generate general formulae for the cardinality of all possible codes for $MM(c,p)$ by appealing to elementary combinatorics. [2] \\
\indent Consider the total number of possible codes as the sum of all of the possible codes containing exactly \textit{i} letters. We denote the number of possibles codes of \textit{p} pegs of exactly \textit{i} colors as $C_{i}$. Each such code for a fixed number of distinct colors \textit{i} amounts to a multinomial coefficient, whereby:
\begin{equation}
\begin{aligned}
|MM(c,p)|=\sum_{i=1}^{min\{c,p\}} C_{i} \\ =\sum_{i=1}^{min\{c,p\}}\binom{c}{i} \sum_{n_1+\ldots +n_i=p}^{} \displaystyle  \binom{p}{n_1,\ldots,n_i} \\ =\sum_{i=1}^{min\{c,p\}}\binom{c}{i} \sum_{n_1+\ldots +n_i=p}^{} \frac{p!}{n_1!\ldots n_i!}
\end{aligned}
\end{equation}
Confirming the above formula for $MM(6,4)$ yields: $|MM(6,4)|=\displaystyle \binom{6}{1}1+\displaystyle \binom{6}{2}\bigg(\frac{4!}{1!3!}+\frac{4!}{2!2!}+\frac{4!}{3!1!}\bigg)+\displaystyle \binom{6}{3}\Big(3\cdot \frac{4!}{1!2!1!} \Big)+\displaystyle \binom{6}{4}4!=6 +15(4+6+4)+20\cdot3\cdot12+15\cdot24\\=1296.$ \\
\indent An early and remarkable result for the mastermind search problem was provided by Knuth [3] which proves that optimally five questions suffices to guarantee a solution to $MM(6,4).$ We now consider a simple generalized lower bound related to this claim. \\
\indent \textbf{Theorem.} For $c\geq 2,$ the minimum number of guesses required to guarantee solution a for $MM(c,2)$, i.e. the general 2-position mastermind game (which we denote $m_{(c,2)}$) is $\floor[\big]{c/2} +2$.[2]\\
\indent \textbf{Proof.} Guess $\ceil[\big]{k/2}$ times using two new colors each turn. From these guesses, the code-breaker can receive a positive response (i.e. a black or white key peg) at most twice. If the code-breaker receives two key pegs of any color in response on two occasions from these queries, one can show that there are a most two possible consistent master codes. Conversely, if the code-breaker receives no positive responses in total, then $c$ must be odd and the master code consists of the lone unused color. In either case, we have: $m_{(c,2)}$ = $ \floor[\big]{c/2} +2$, as desired. 
\indent It therefore follows follows that $\forall c,p \geq 2$, $m_{(c,p)} \geq$ $\floor[\big]{c/2} +2$, yielding a general lower bound. In particular, for Knuth's result, the bound is tight, as $m_{(6,4)}$ =$\floor[\big]{6/2} +2=5.$\\
\indent Bondt [11] showed that solving a mastermind board for $M(c,2)$ is nevertheless an NP-complete problem, by reducing it to the \textit{3SAT problem}. Moreover, the 
\textit{mastermind satisfiability problem} which asks, given a set of queries and corresponding key peg replies, whether there exists a master code that satisfies this set of query-key conditions, has been shown to be NP-complete [12]. \\ 
\indent A broad range of search algorithms have been previously applied to mastermind and its variants. Knuth's method from 1977 (detailed in the next section) applies a minimax search using a heuristic based on the size of query-partitions. This method yielded 4.467 expected queries, with a maximum of five queries for all possible codes in $MM(6,4)$. Of note, the $MM(6,4)$ variation of mastermind was effectively solved in 1993 by Koyama and Lai using exhaustive, depth-first search achieving 4.34 expected queries -- the search time \textit{per puzzle} at publication was, however, on the order of several hours. \\
\indent Beyond complete search, genetic algorithms (GAs) have been applied extensively to mastermind, including [6], [7] and [10]. In the GA paradigm, a large set of "eligible" codes (e.g. consistent codes -- although several approaches show that inconsistent codes are sometimes more informative for mastermind search) is considered for each generation. The "goodness" of these codes is determined using a fitness function which assigns a score to each code based on its probability of being the master code using various meta-heuristics. At each generation, standard genetic operations including crossover, mutation and inversion are applied in order to render the new population. [7] In particular achieved 4.39 expected queries for $MM(6,4)$ using a fitness function defined by a weighted sum of L1 key peg differences between candidate codes and query codes. \\
\indent Shapiro [13] adopts a method which simply draws queries from the set of codes consistent with all previous queries; Blake et al [5] use an MCMC approach; [6] combine hill-climbing and heuristics, while Cover and Thomas [14] introduce an information theoretical strategy. \\
\indent In the current work we apply Simulated Annealing to the mastermind problem in 
addition to introducing a novel search heuristic which aims to maximize the expected reduction in the set of codes consistent with the master code. 
\section{Knuth's Method}
\indent Knuth's method, the "five guess algorithm" for $MM(6,4)$ works as follows. The first guess is deterministically chosen as 1122 (Knuth provides examples showing that beginning with a different choice such as 1123 or 1234 can lead to situations where more than five queries are required to solve the puzzle). Following this initial guess, the code-maker responds with the corresponding key pegs, and using this response, the code-breaker generates a set of consistent codes $\textit{S}.$ Next, the code-breaker applies a minimax technique to the set \textit{S}, where each node in the search tree is evaluated according to the expected size of the various $Q_p$ query-partitions; in particular, the evaluation function returns the size of the maximum partition for each consistent code, and the code with the maximum partition of (expected) minimal size is chosen as the next query. This process is repeated until termination.

\begin{algorithm}[h]
\caption{Knuth's Method}
\begin{algorithmic}
\STATE Code-breaker sets initial query code: $q_{1} \leftarrow 1122$.
\STATE Code-maker replies with key pegs: $[b_{1},w_{1}]$;
\WHILE{$b_i$ != 4}
  \STATE Generate consistency set $S_i$
  \STATE Compute expected size of maximum query-partition for each code in $S$: $E_{c \in S} [Max|Q|]$
  \STATE $q_i \leftarrow Min(E_{c \in S} [Max(|Q|]))$
  \STATE Code-maker replies with key pegs: $[b_{i},w_{i}]$;
  \ENDWHILE
\end{algorithmic}
\label{alrc_algorithm}
\end{algorithm}
\indent To further illustrate Knuth's method, Table II explicitly shows the size of each query-partition for $MM(6,4)$ (recall that $Q_{p}=14$) for generic initial query types: $1111$, $1112$, $1122$, $1123$, and $1234$. Note that according to Knuth's criterion, $1122$ would be chosen in this case because it yields the smallest of all maximum partition sets of the given queries. 

\begin{table}
\centering
\begin{tabular}{|l|c|c|c|c|c|}\hline
& 1111 & 1112 & 1122 & 1123 & 1234 \\ \hline
[0,0] & 625 & 256 & 256 & 81 & 16 \\ \hline
[0,1] & 0 & 308 & 256 & 276 & 152 \\ \hline
[0,2] & 0 & 61 & 96 & 222 & 312 \\ \hline
[0,3] & 0& 0 & 16 & 44 & 136 \\ \hline
[0,4] & 0 & 0 & 1 & 2 & 9 \\ \hline
[1,0] & 500 & 317 & 256 & 182 & 108 \\ \hline
[1,1] & 0 & 156 & 208 & 230 & 252 \\ \hline
[1,2] & 0 & 27 & 36 & 84 & 132 \\ \hline
[1,3] & 0 & 0 & 0 & 4 & 8 \\ \hline
[2,0] & 150 & 123 & 114 & 105 & 96 \\ \hline
[2,1] & 0 & 24 & 32 & 40 & 48 \\ \hline
[2,2] & 0 & 3 & 4 & 5 & 6 \\ \hline
[3,0] & 20 & 20 & 20 & 20 & 20 \\ \hline
[4,0] & 1 & 1 & 1 & 1 & 1 \\ \hline
\end{tabular}
\caption{Enumeration of all query-partition sizes for various initial codes: 1111, 1112, 1122, 1123, 1234. Based on the heuristic used in Knuth's method, code 1122 is chosen because it generated the smallest maximum size partition.} 
\end{table}

\section{Simulated Annealing}
We apply Simulated Annealing (SA) to the mastermind search problem. More concretely, our method combines elements of stochastic, local hill-climbing (a la SA) with non-local, consistency-based search. At each step $i$ of the algorithm we construct the the consistency set $S_i$ comprising the subset of codes consistent with all previous queries. We augment this set with a "neighborhood" consisting of the set of codes with Hamming distance one from the current query code $c_i$ (observe that these codes need not be consistent with the given queries) to form the set $S_i'$. We score these neighbors using the following scoring function: 
\begin{equation}
score(c)=\sum_{q \in \text{queries}}^{}|q_{b}(c)-q_{b}(q)|+ |q_{w}(c)-q_{w}(q)|
\end{equation}
Where $q_b(\cdot)$ and $q_w(\cdot)$ indicate the number of black and white key pegs generated for a given query code, respectively; observe that $score(c) = 0 \text{ iff } c \in S_i$ Finally, we randomly sample codes from the augmented set $S_i'$. If the code $c$ is consistent with the query histories it is automatically chosen as the next query; otherwise, it is accepted as the next query with acceptance probability: $\frac{\alpha}{score(c)+1}$ (we use $\alpha = 2$ in experiments). This acceptance probability formula encodes an implicit "temperature", since the acceptance property of inconsistent codes decreases in proportion to the algorithm step number -- which is to say the algorithm "anneals" over time. We gives a pseudo-code description of our SA procedure below. 
\begin{algorithm}[h]
\caption{Simulated Annealing}
\begin{algorithmic}
\STATE Code-breaker sets initial query code $c_1$ randomly.
\STATE Code-maker replies with key pegs: $[b_{1},w_{1}]$;
\WHILE{$b_i$ != 4}
  \STATE Generate consistency set $S_i$
  \STATE Generate neighborhood of $c_i$, $N_{c_i}$ using Hamming distance of one
  \STATE Form augmented set: $S_i'=S_i \cup N_{c_i}$
  \STATE Sample random code $c \in S_i'$ 
  \IF{$c \in S_i$} \STATE{$q_i \leftarrow c$}
  \ELSE \STATE{with probability $\frac{\alpha}{score(c)+1}$, $q_i \leftarrow c$}
  \STATE {where $score(c)=\sum_{q \in \text{queries}}^{}|q_{b}(c)-q_{b}(q)|+$}
  \STATE {$|q_{w}(c)-q_{w}(q)|$}
  \ENDIF
  \ENDWHILE
\end{algorithmic}
\label{alrc_algorithm}
\end{algorithm}

\section{Maximum Expected Reduction in Consistency}
\indent We introduce a novel search heuristic defined as the expected reduction in the size of the consistent code set for mastermind search. At each step $i$ of the MERC algorithm, we first determine the expected reduction in the size of the consistency set $S_i$ for each code $c \in S_i$. Concretely, $\forall c \in S_i$ we generate responses over all possible candidate master codes $m' \in S_i: [b_{c,m'},w_{c,m'}]=q(c,m').$ Next, for each such $m'$ we compute $q(m',c')$ and count the size of the set of codes $c' \in S_i$ such that $q(m',c')=q(c,m')$. The cardinality of this set represents the expected size of the consistency set with respect to code $c$. By choosing the code $c$ for which the cardinality of the expected size of the consistency set is smallest, we render the maximum expected reduction in the consistency set. 
\begin{algorithm}[h]
\caption{Maximum Expected Reduction in Consistency (MERC)}
\begin{algorithmic}
\STATE Code-breaker sets initial query code: $q_{1} \leftarrow 1122$.
\STATE Code-maker replies with key pegs: $[b_{1},w_{1}]$;
\WHILE{$b_i$ != 4}
  \STATE Generate consistency set $S_i$
  \FOR{$c,m' \in S_i$} 
  \STATE generate responses over all possible candidate master codes $m' \in S_i: [b_{c,m'},w_{c,m'}]=q(c,m')$
     \FOR {$ c' \in S_i$} 
     \STATE compute $q(m',c')$ -- with $m'$ fixed -- and count the size of the set of codes $c' \in S_i$ such that $q(m',c')=q(c,m')$
        \ENDFOR
    \ENDFOR
    \STATE for c for which the cardinality of the expected size of the consistency set is smallest: $q_{i} \leftarrow c$
  \ENDWHILE
\end{algorithmic}
\label{alrc_algorithm}
\end{algorithm}

\section{Experimental Results}
\indent We generated experimental results using simulations of $n=5000$ games for the $MM(6,4)$, $MM(5,4)$ and $MM(4,4)$ variants of mastermind. In addition to Knuth's method, the SA algorithm and MERC algorithm described previously, we also implemented a baseline uninformed random search algorithm. For each experiment, we report the mean, median, standard deviation and maximum number of queries required. The Knuth and MERC algorithms used a deterministic choice of initial code (1122), while the random and SA algorithms used a random initial code. Our experimental results are summarized in Table III. 
\begin{table}
\centering
\begin{tabular}{|l|p{19mm}|p{15mm}|p{15mm}|}\hline
Algorithm & MM(6,4) & MM(5,4) & MM(4,4) \\ \hline
Random & mean: 639.9 \newline Max: 1296 \newline Median: 634.5 \newline STD: 370.8  & 
mean: 315.1 \newline Max: 625 \newline Median: 315 \newline STD: 180.2 &
mean: 131.2 \newline Max: 256 \newline Median: 132  \newline STD: 74.13 \\ \hline
Knuth & mean: 4.468 \newline Max: 7 \newline Median: 5.0 \newline STD: .7322 &
mean: 4.105 \newline Max: 6 \newline Median: 4.0 \newline STD: .7321 & 
mean: 3.631 \newline Max: 5 \newline Median: 4.0 \newline STD: .6843 \\ \hline
SA & mean: 5.7916 \newline Max: 13 \newline Median: 6.0 \newline STD: 1.673 &
mean: 5.1306 \newline Max: 12 \newline Median: 5.0 \newline STD: 1.488 &
mean: 4.3826 \newline Max: 11 \newline Median: 4.0 \newline STD: 1.229 \\ \hline
MERC & mean: 4.714 \newline Max: 7 \newline Median: 5.0 \newline STD: .8954 & 
mean: 4.206 \newline Max: 7 \newline Median: 4.0 \newline STD: .8472 &
mean: 3.751 \newline Max: 6 \newline Median: 4.0 \newline STD: .820 \\ \hline

\end{tabular}
\caption{Experimetal results summary for Random, Knuth, SA and MERC algorithms applied to $MM(6,4)$,$MM(5,4)$ and $MM(4,4)$ using $n=5000$ sample games.} 
\end{table}
\section{References} 
[1] Goddard, Wayne. Mastermind Revisited. 2004.

[2] Ville, Geoffroy. An Optimal Mastermind (4,7) Strategy and More Results in the Expected Case, 2013. 

[3] Knuth, D.E. The Computer as Mastermind. Journal of Recreational Mathematics, 1976-77, 1–6.

[4] Peter Norvig. 1984. Playing Mastermind optimally. SIGART Bull. 90 (October 1984), 33-34. 

[5] Blair, Nathan, et al. Mastering Mastermind with MCMC. 2018. 

[6] Temporel, Alexandre, et al. A Heuristic Hill-Climbing Algorithm for Mastermind. 2004.

[7] Berghman, Lotte, et al. Efficient Solutions for Mastermind Using Genetic Algorithms. Computers and Operations Research, Volume 36, Issue 6, June 2009.

[8] Kooi, Barteld.Yet another Mastermind strategy. ICGA Journal. 28. 10.3233/ICG-2005-28105. 2005.

[9] Singley, Andrew. Heuristic Solution Methods for the 1-Dimensional and the 2-Dimensional Mastermind Problem. 2005 (MS thesis).

[10] Kalisker, Tom, et al. Solving Mastermind Using Genetic Algorithms. GECCO 2003, LNCS 2724. 2003. 

[11] Bondt, Michiel. NP-completeness of Master Mind and Minesweeper. Journal of Physical Chemistry A - J PHYS CHEM A. 2004.

[12] Stuckman, J., and Zhang, G. Mastermind is NP-Complete. ArXiv. 2005.

[13] Shapiro, Ehud.  Playing Mastermind Logically. SIGART Bull. 85 (July 1983).

[14] Cover, Thomas and Joy A. Thomas.Elements of Information Theory (Wiley Series in Telecommunications and Signal Processing). Wiley-Interscience, New York, NY, USA. 2006.

\clearpage

\end{document}